**Title**

M³Fair: Mitigating Bias in Healthcare Data through Multi-Level and Multi-Sensitive-Attribute Reweighting Method

**Authors**

Yinghao Zhu[1,2]†, Jingkun An[2,3]†, Enshen Zhou[2,3], Lu An[4], Junyi Gao[5,6], Hao Li[2,3], Haoran Feng[2,3], Bo Hou[2,3], Wen Tang[7], Chengwei Pan[1*], Liantao Ma[2*]

**Affiliations**

[1]*Institute of Artificial Intelligence, Beihang University, Beijing, China*
[2]*National Engineering Research Center for Software Engineering, Peking University, Beijing, China*
[3]*School of Software, Beihang University, Beijing, China*
[4]*Department of Electrical and Computer Engineering, North Carolina State University, North Carolina, USA*
[5]*Institute of Genetics and Cancer, University of Edinburgh, Edinburgh, UK*
[6]*Health Data Research UK, UK*
[7]*Department of Nephrology, Peking University Third Hospital, Beijing, China*

*Address correspondence to: pancw@buaa.edu.cn (C.P.); malt@pku.edu.cn (L.M.)
†These authors contributed equally to this work.

## 1. Background

In the data-driven artificial intelligence paradigm, models heavily rely on large amounts of training data. However, factors like sampling distribution imbalance can lead to issues of bias and unfairness in healthcare data. Sensitive attributes, such as race, gender, age, and medical condition, are characteristics of individuals that are commonly associated with discrimination or bias. In healthcare AI, these attributes can play a significant role in determining the quality of care that individuals receive. For example, minority groups often receive fewer procedures and poorer-quality medical care than white individuals in US [1]. Therefore, detecting and mitigating bias in data is crucial to enhancing health equity.

Bias mitigation methods include pre-processing, in-processing, and post-processing [2]. Among them, Reweighting (RW) is a widely used pre-processing method that performs well in balancing machine learning performance and fairness performance [2]. RW adjusts the weights for samples within each (group, label) combination, where these weights are utilized in loss functions [3]. However, RW is limited to considering only a single sensitive attribute when mitigating bias and assumes that each sensitive attribute is equally important. This may result in potential inaccuracies when addressing intersectional bias [4].

To address these limitations, we propose M³Fair, a multi-level and multi-sensitive-attribute reweighting method by extending the RW method to multiple sensitive attributes at multiple levels. Our experiments on real-world datasets show that the approach is effective, straightforward, and generalizable in addressing the healthcare fairness issues.

## 2. Objectives

(1) Detect and identify source(s) of biases. There may be multiple features that could potentially lead to bias. These features are defined as sensitive attributes.

(2) Mitigate bias impact of sensitive attributes while balancing machine learning and fairness performance, ensuring model effectiveness, accuracy, and equity in healthcare AI.

## 3. Methods

### 3.1. Bias Detection

To identify sensitive attributes, we consider features that consistently exhibit biased tendencies across various bias evaluation metrics as sensitive attributes. Concretely, we first binarize each feature by comparing its mean value. Then, iterate through each feature, calculating its performance on four evaluation metrics: Disparate Impact (DI), Statistical Parity Difference (SPD), Average Odds Difference (AOD), and Equal Opportunity Difference (EOD) [2]. These metrics reflect different aspects of bias as they capture different ways in which privileged groups can be treated unfairly [5], and their combined use allows us to identify features that consistently exhibit bias across multiple dimensions. Thus, we take the intersection of the top $N$ most unfair features based on these metrics (default $N = 20$). Our experiment results show that features such as race and age consistently exhibit bias across these metrics, highlighting the importance of identifying and mitigating bias based on multiple sensitive attributes simultaneously.

### 3.2. Bias Mitigation

We first define sensitivity levels (SL) as the sum of level weights for multiple sensitive attributes associated with a sample, allowing us to assign level weights to each sensitive attribute. Next, M³Fair calculates the sum of level weights for samples with or without favorable labels. The weight coefficients for samples are then computed based on Equation (1). through iterating each sensitivity level.

$$W'_{i, y_i=d} = W_i \frac{\sum_{y_j=d} W_j \cdot \sum_{SL} W_j}{\sum W_j \cdot \sum_{y_j=d, SL} W_j} \quad (1)$$

Where $d \in \{0,1\}$, $d = 1$ means the sample has a favorable label (typically class 1) and $d = 0$ means it has an unfavorable label (typically class 0), $W$ is the sample weight. The calculated sample weights $W'$ can be simply applied in the loss function of models.

## 4. Results

**Table 1** shows our experiment results. SA represents sensitive attributes. EA denotes the evaluated attribute. A desirable score for DI is the score close to 1, while SPD, AOD, and EOD metrics are the lower, the better. The notation A->B requires reweighting twice, first by attribute A and then by B, while the notation [A, B] means A and B attributes are reweighted simultaneously. The model used is a logistic regression model. We choose the sensitivity level setting of Sex=1, Race=2, Age=2 in the M$^3$Fair method through grid search in the search space of {1, 2}.

**Table 1. Experiment results on Adult, TJH, CDSL datasets.**

| Data | Method | SA | EA | ACC | AUROC | AUPRC | DI | SPD | AOD | EOD |
|---|---|---|---|---|---|---|---|---|---|---|
| Adult | / | / | Sex | 0.8007 | 0.8216 | 0.7300 | 0.3635 | -0.1989 | -0.2089 | -0.1672 |
| | | | Race | | | | 0.6038 | -0.1040 | -0.0886 | -0.0604 |
| | RW | Sex->Race | Sex | 0.7856 | 0.8138 | 0.7216 | 0.9786 | 0.0534 | -0.0583 | 0.0111 |
| | | | Race | | | | 1.0000 | 0.0000 | -0.0112 | 0.0223 |
| | | Race->Sex | Sex | 0.7857 | 0.8141 | 0.7219 | 1.0000 | 0.0000 | -0.0631 | 0.0081 |
| | | | Race | | | | 0.913 | 0.0232 | 0.0069 | 0.0356 |
| | M$^3$Fair | [Sex, Race] | Sex | 0.7858 | 0.8143 | 0.7221 | **1.0000** | **0.0000** | -0.0627 | **0.0078** |
| | | | Race | | | | **1.0000** | **0.0000** | **-0.0106** | **0.0214** |
| TJH | / | / | Sex | 0.8349 | 0.8357 | 0.8696 | 0.5454 | -0.2573 | -0.0699 | -0.0018 |
| | | | Age | | | | 0.2782 | -0.4824 | -0.2585 | -0.2922 |
| | RW | Sex->Age | Sex | 0.8073 | 0.8095 | 0.8500 | 0.9789 | 0.0098 | -0.0392 | -0.0281 |
| | | | Age | | | | 1.0000 | 0.0000 | -0.2191 | -0.3160 |
| | | Age->Sex | Sex | 0.8073 | 0.8095 | 0.8500 | 1.0000 | 0.0000 | -0.0392 | -0.0281 |
| | | | Age | | | | 0.9891 | 0.0050 | -0.2191 | -0.3160 |
| | M$^3$Fair | [Sex, Age] | Sex | 0.8073 | 0.8095 | 0.8500 | **1.0000** | **0.0000** | **-0.0392** | -0.0281 |
| | | | Age | | | | **1.0000** | **0.0000** | **-0.2191** | -0.3160 |
| CDSL | / | / | Sex | 0.6766 | 0.7254 | 0.5397 | 0.1374 | -0.1858 | -0.5453 | -0.3878 |
| | | | Age | | | | 0.8368 | -0.0222 | -0.1077 | -0.2002 |
| | RW | Sex->Age | Sex | 0.7776 | 0.6497 | 0.4175 | 0.8986 | 0.0134 | -0.0677 | -0.0685 |
| | | | Age | | | | 1.0000 | 0.0000 | -0.0483 | 0.0270 |
| | | Age->Sex | Sex | 0.8081 | 0.6547 | 0.4248 | 1.0000 | 0.0000 | -0.0084 | -0.0170 |
| | | | Age | | | | 0.9806 | -0.0025 | -0.0354 | 0.0608 |
| | M$^3$Fair | [Sex, Age] | Sex | 0.8121 | 0.6569 | 0.4285 | **1.0000** | **0.0000** | **-0.0027** | **-0.0170** |
| | | | Age | | | | **1.0000** | **0.0000** | **-0.0268** | 0.0608 |

We conducted experiments on the UCI Adult dataset and two real-world COVID-19 EHR datasets, TJH and CDSL [6]. As shown in table 1, compared to not using any mitigation methods, our M$^3$Fair method successfully mitigates biases introduced by sensitive attributes such as gender, age, and race in 91.67% (22/24) metrics, while slightly affecting model performance with an average performance drop of 2.51%. Furthermore, M$^3$Fair shows 100% better (12 metrics) or equal (12 metrics) fairness performance than the baseline Reweighting method on all four metrics.

## 5. Conclusions

The proposed method M³Fair focuses on the bias within the dataset. We detect sensitive attributes by analyzing consistently biased features. By extending the existing Reweighting method, we support multiple sensitivity levels and sensitive attributes. Experiments demonstrate that M³Fair with considering multiple sensitive attributes successfully mitigates bias in 91.67% metrics and performs better or equivalent than a single attribute in all metrics. Our method is generalizable and scalable, allowing for customization of sensitive levels in clinical decision-making processes. The code is open-sourced at https://github.com/yhzhu99/M3Fair.

Our solution won the NIH Bias Detection Tools for Clinical Decision Making Challenge and was evaluated by NIH as "performed well in the test harness and had a relatively strong performance across social and predictive fairness metrics. The approach is straightforward, and very generalizable to any number of use-cases."